\documentclass[sigconf,balance=false,authorversion=true,screen=true]{acmart}

\renewcommand\footnotetextcopyrightpermission[1]{}
\renewcommand\acmConference[1]{}{}{}

\usepackage{graphicx}
\usepackage{amsmath}
\usepackage{booktabs}
\usepackage{caption}
\usepackage{subcaption}
\usepackage{multirow}
\usepackage{pifont}
\usepackage[prologue]{xcolor} 
\usepackage{colortbl}
\usepackage{soul}
\usepackage{listings}
\usepackage{enumitem}
\usepackage{xspace}

\definecolor{yesColor}{RGB}{37,37,37}
\definecolor{noColor}{RGB}{189,189,189}
\newcommand{\cmark}{\textcolor{yesColor}{\ding{51}}}%
\newcommand{\xmark}{\textcolor{noColor}{\ding{55}}}%
\newcommand{\repo}{\texttt{CVNets}\xspace}

\definecolor{codegreen}{rgb}{0,0.6,0}
\definecolor{codegray}{rgb}{0.5,0.5,0.5}
\definecolor{codepurple}{rgb}{0.58,0,0.82}
\definecolor{backcolour}{rgb}{0.95,0.95,0.92}

\definecolor{highlightcolor}{RGB}{255,255,191}

\lstdefinestyle{mystyle}{
    backgroundcolor=\color{backcolour},   
    commentstyle=\color{codegreen},
    keywordstyle=\color{magenta},
    numberstyle=\tiny\color{codegray},
    stringstyle=\color{codepurple},
    basicstyle=\ttfamily\footnotesize,
    breakatwhitespace=false,         
    breaklines=true,                 
    captionpos=b,                    
    keepspaces=true,                 
    numbers=left,                    
    numbersep=5pt,                  
    showspaces=false,                
    showstringspaces=false,
    showtabs=false,                  
    tabsize=2
}

\DeclareRobustCommand{\hlyellow}[1]{{\sethlcolor{highlightcolor}\hl{#1}}}

\AtBeginDocument{%
  \providecommand\BibTeX{{%
    \normalfont B\kern-0.5em{\scshape i\kern-0.25em b}\kern-0.8em\TeX}}}

\begin{document}

\title{\repo: High Performance Library for Computer Vision}

\author{Sachin Mehta}
\authornote{Project lead and main contributor}
\affiliation{%
  \institution{Apple}
   \country{}
}

\author{ Farzad Abdolhosseini}
\affiliation{%
  \institution{Apple}
   \country{}
}

\author{Mohammad Rastegari}
\affiliation{%
  \institution{Apple}
   \country{}
}

\renewcommand{\shortauthors}{Mehta, et al.}

\begin{abstract}
  We introduce \repo, a high-performance open-source library for training deep neural networks for visual recognition tasks, including classification, detection, and segmentation. \repo~supports image and video understanding tools, including data loading, data transformations, novel data sampling methods, and implementations of several standard networks with similar or better performance than previous studies. Our source code is available at: \url{https://github.com/apple/ml-cvnets}.
\end{abstract}



\begin{teaserfigure}
    \centering
    \begin{subfigure}[b]{0.45\columnwidth}
        \centering
        \includegraphics[width=0.85\columnwidth]{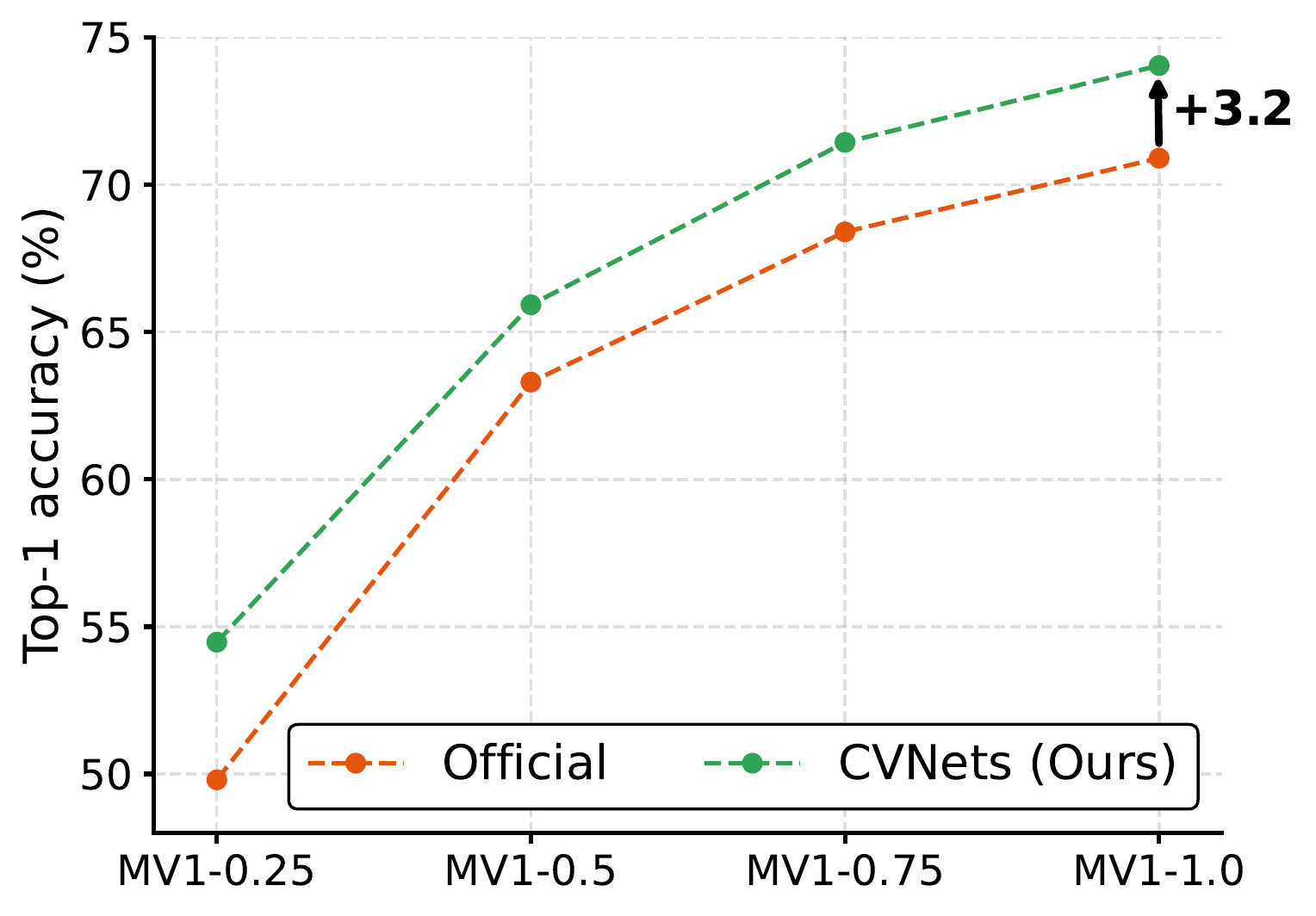}
        \caption{MobileNetv1 (MV1) at different width factors}
    \end{subfigure}
    \hfill
    \begin{subfigure}[b]{0.45\columnwidth}
        \centering
        \includegraphics[width=0.85\columnwidth]{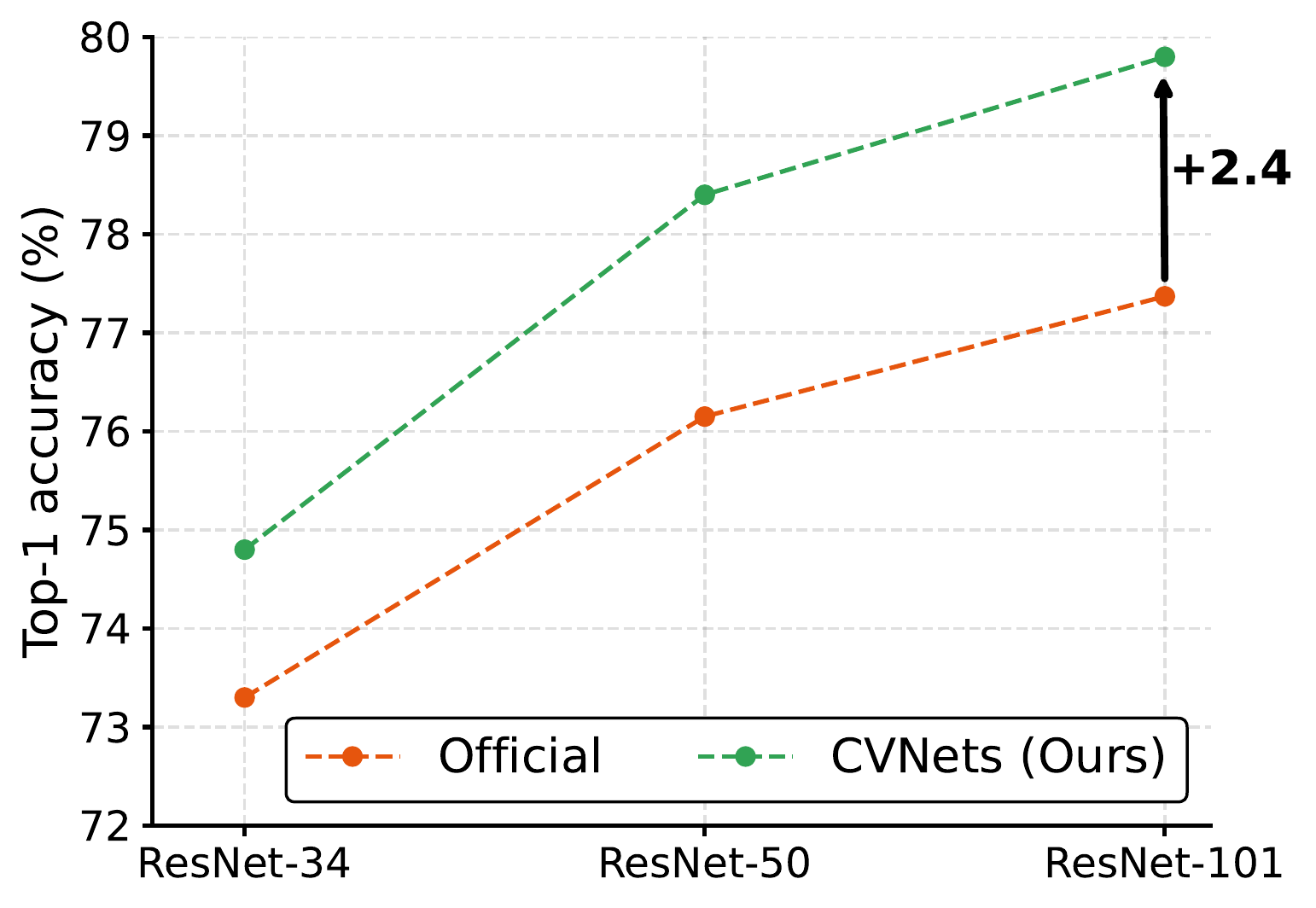}
        \caption{ ResNet at different depths}
    \end{subfigure}
  \caption{\repo~ can be used to improve the performance of different deep neural networks on the ImageNet dataset significantly with simple training recipes (e.g., random resized cropping and horizontal flipping). The official MobileNetv1 \cite{howard2017mobilenets} and ResNet \cite{he2016deep} results are from TensorflowLite \cite{tensorflow2015whitepaper} and Torchvision \cite{pytorch2022pytorch}, respectively.}
  \label{fig:teaser}
\end{teaserfigure}

\maketitle

\section{Introduction}
With the rise of deep learning, significant progress has been made in visual understanding tasks, including novel light- and heavy-weight architectures, dedicated hardware and software stacks, advanced data augmentation methods, and better training recipes. There exist several popular libraries that provide implementations for different tasks and input modalities, including Torchvision \cite{pytorch2022pytorch}, TensorflowLite \cite{tensorflow2015whitepaper}, \texttt{timm} \cite{wightman2021resnet}, and PyTorchVideo \cite{fan2021pytorchvideo}. Many of these libraries are modular and are designed around a particular task and input modality, and provide implementations and pre-trained weights of different networks with varying performance. However, \emph{reproducibility varies across these libraries}. For example, Torchvision library uses advanced training recipes (e.g., better augmentation) to achieve the same performance for training MobileNetv3 on the ImageNet dataset \cite{deng2009imagenet} as TensorflowLite with simple training recipes. 

We introduce \repo, a PyTorch-based deep learning library for training computer vision models with higher performance. With \repo, we enable researchers and practitioners in academia and industry to train either novel or existing deep learning architectures with high-performance across different tasks and input modalities. \repo is a modular and flexible framework that aims to train deep neural networks faster with simple or advanced training recipes. Simple recipes are useful for research in resource-constrained environments as they train models for fewer epochs with  basic data augmentation (random resized crop and flipping) as compared to advanced training recipes, which trains model for $2-4 \times$ longer with advanced augmentation methods (e.g., CutMix and MixUp). With simple recipes (similar to the ones in original publications) and variable batch sampler (Section \ref{ssec:data_samplers}), \repo~improves the performance of ResNet-101 significantly (Figure \ref{fig:teaser}) on the ImageNet dataset while for advanced training recipes with the same batch size and number of epochs, it delivers similar performance to previous methods while requiring $1.3 \times$ fewer optimization updates.

\section{\repo~Library Design}
\repo~follows the design principles below:

\paragraph{\bfseries Modularity} \repo~provides independent components; allowing users to plug-and-play different components across different visual recognition tasks for both research and production use cases. \repo~ implement different components, including datasets and models for different tasks and input modalities, independently. For example, different classification backbones (e.g., ResNet-50) trained in \repo~can be seamlessly integrated with object detection (e.g., SSD) or semantic segmentation (e.g., DeepLabv3) pipelines for studying the generic nature of an architecture.

\paragraph{\bfseries Flexibility} With \repo, we would like to enable new use cases in research as well as production. We designed \repo~such that new components (e.g., models, datasets, loss functions, data samplers, and optimizers) can be integrated easily. We achieve this by registering each component. As an example, ADE20k dataset for the task of segmentation is registered in \repo~as:
\hlyellow{\texttt{\\@register\_dataset(name=``ade20k'', task=``segmentation'')}}

To use this dataset for training, one can use \texttt{dataset.name} and \texttt{dataset.category} as command line arguments. 

\paragraph{\bfseries Reproducibility} \repo~provide reproducible implementations of standard models for different computer vision tasks. Each model is benchmarked against the performance reported in original publications as well as the previous best reproduction studies. The pre-trained weights of each model are released online to enable future research.

\paragraph{\bfseries Compatibility} \repo~is compatible with hardware accelerated frameworks (e.g., CoreML) and domain-specific libraries (e.g., PyTorchVideo). The models from domain-specific libraries can be easily consumed in the \repo, as shown in Listing \ref{lst:compatibility_example}; reducing researchers overhead in implementing new components or sub-modules in \repo.
\begin{lstlisting}[frame=tb, xleftmargin=6mm, xrightmargin=6mm, language=Python, caption={An example of registering a video classification model from PyTorchVideo inside CVNets on the Kinetics-400 dataset}, captionpos=b, label={lst:compatibility_example}]
from pytorchvideo.models import resnet 

@register_video_cls_models("resnet_3d")
class ResNet3d(BaseVideoEncoder):
    def __init__(self, opts):
        super().__init__(opts=opts)
        self.model = resnet.create_resnet(
            input_channel=3,
            model_depth=50,
            model_num_class=400,
            norm=nn.BatchNorm3d,
            activation=nn.ReLU,
        )
\end{lstlisting}

\paragraph{\bfseries Beyond ImageNet} Many standard models are benchmarked on the ImageNet dataset. With \repo, we would like to enable researchers to build generic computer vision architectures that can be easily scaled to down-stream tasks such as segmentation and detection. Any classification backbone in \repo~(either existing or new) can seamlessly be integrated with down-stream networks (e.g., PSPNet and SSD) and enables researchers to study the generic nature of different classification models.

\begin{figure*}[t!]
    \centering
    \begin{subfigure}[b]{2\columnwidth}
        \centering
        \resizebox{\columnwidth}{!}{
            \begin{tabular}{cp{0.1cm}cp{0.1cm}c}
               \includegraphics[height=80px]{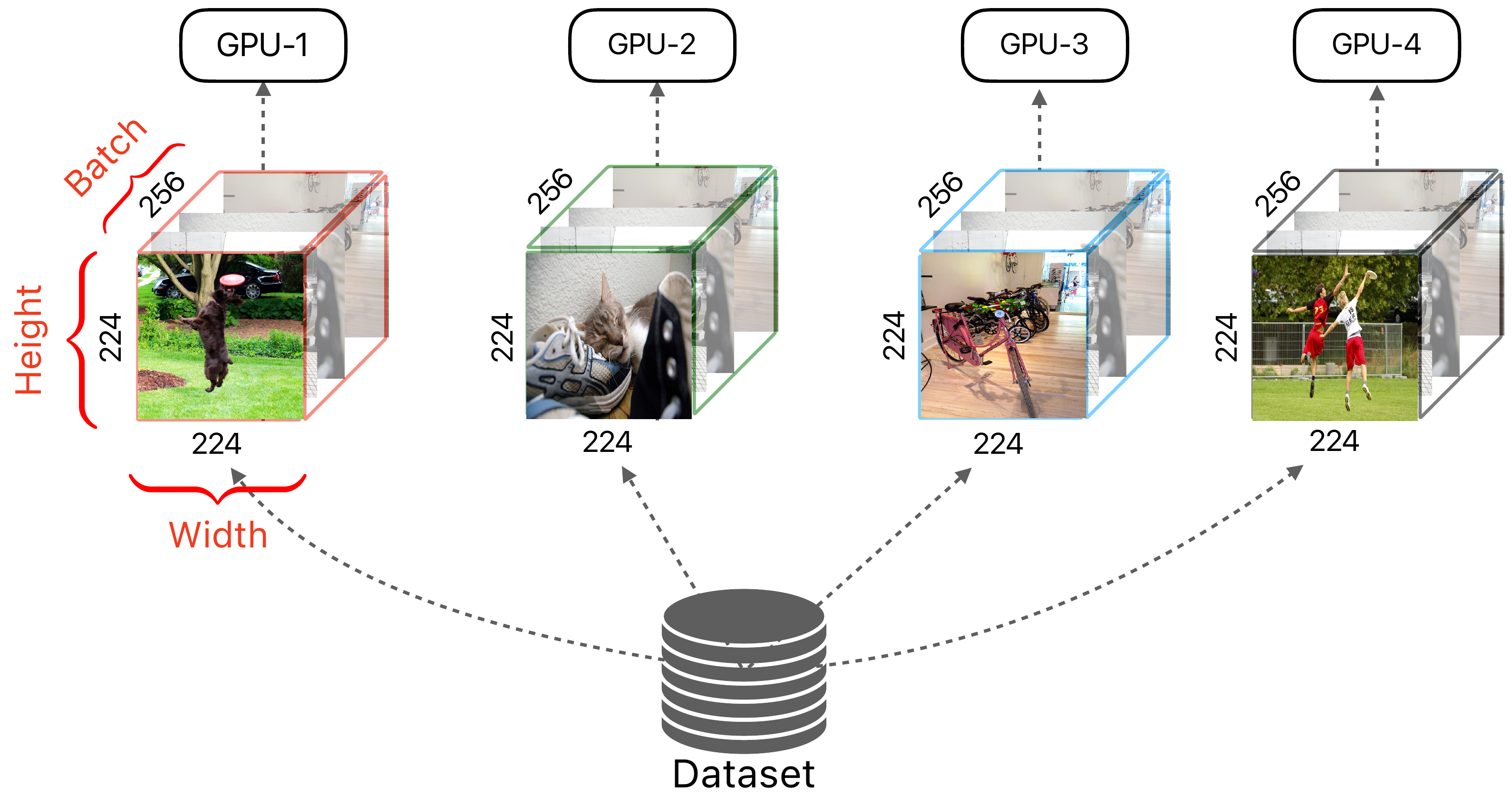}  && \includegraphics[height=80px]{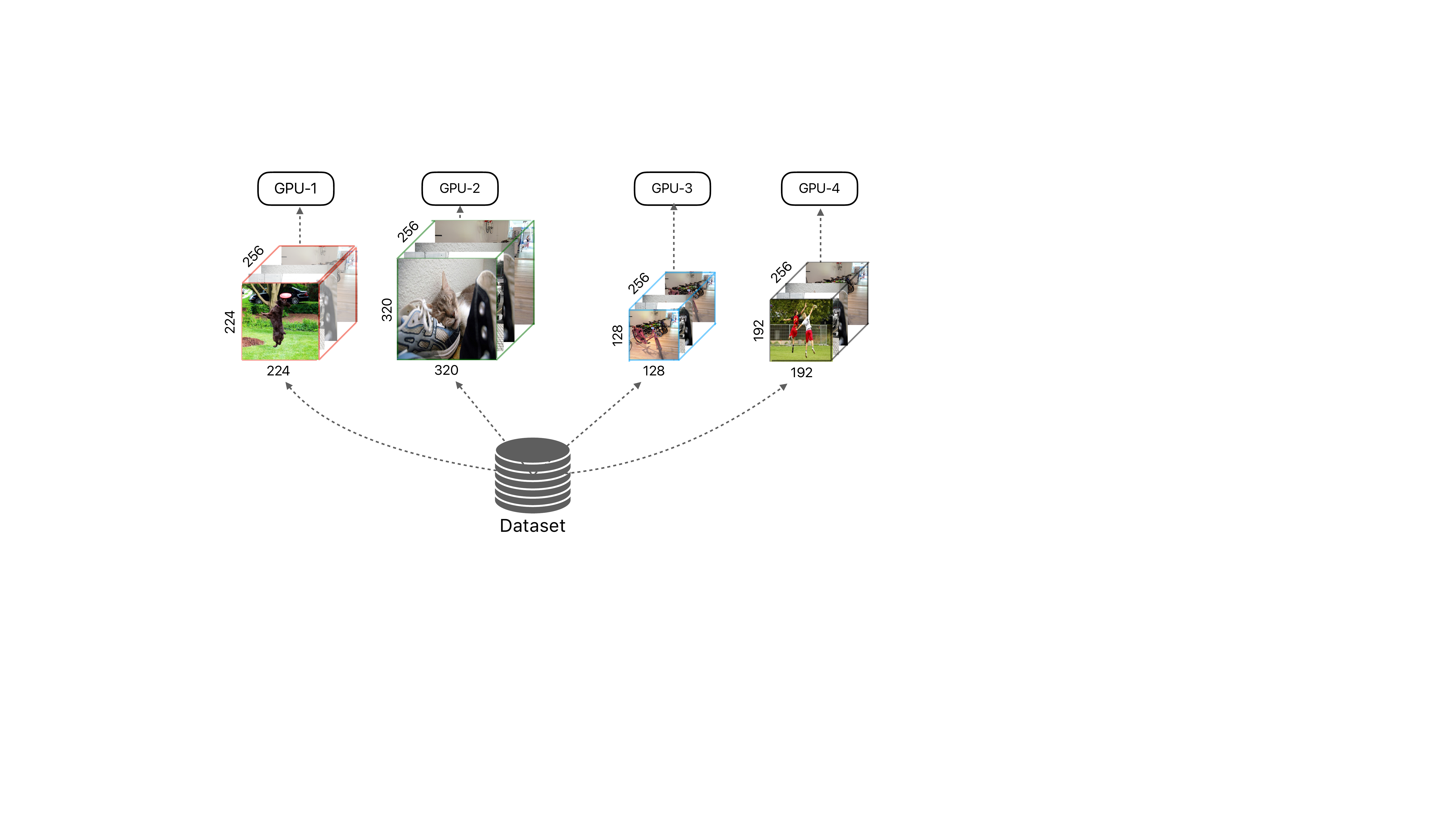} && \includegraphics[height=80px]{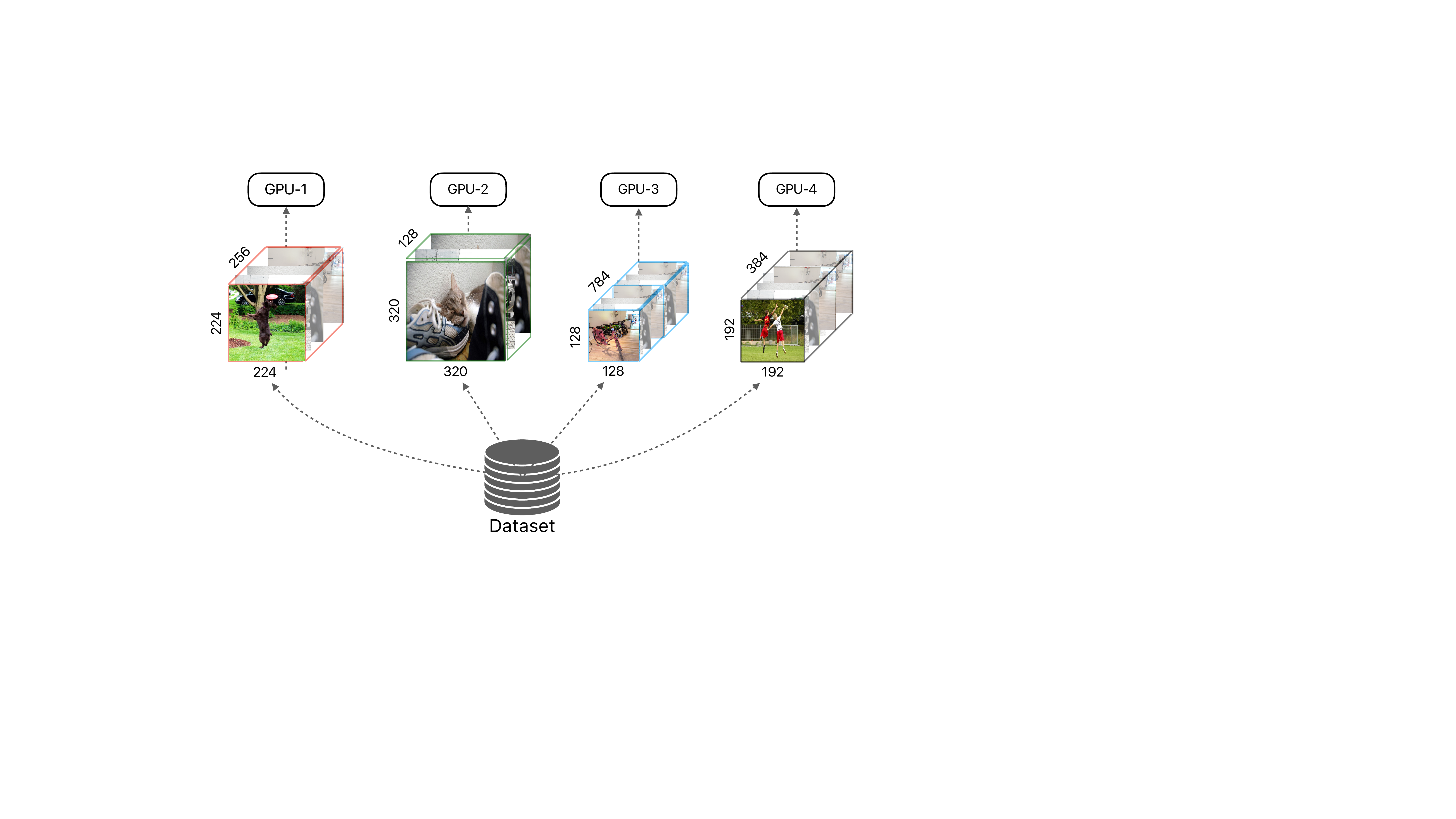} \\
               Single-scale fixed batch size (SSc-FBS) && Multi-scale fixed batch size (MSc-FBS) && Multi-scale variable batch size (MSc-VBS) \\
            \end{tabular}
        }
        \caption{Different samplers}
        \label{fig:data_samplers}
    \end{subfigure}
    \vfill
    \begin{subfigure}[b]{2\columnwidth}
        \centering
        \begin{tabular}{ccp{0.2cm}cc}
             \includegraphics[height=80px]{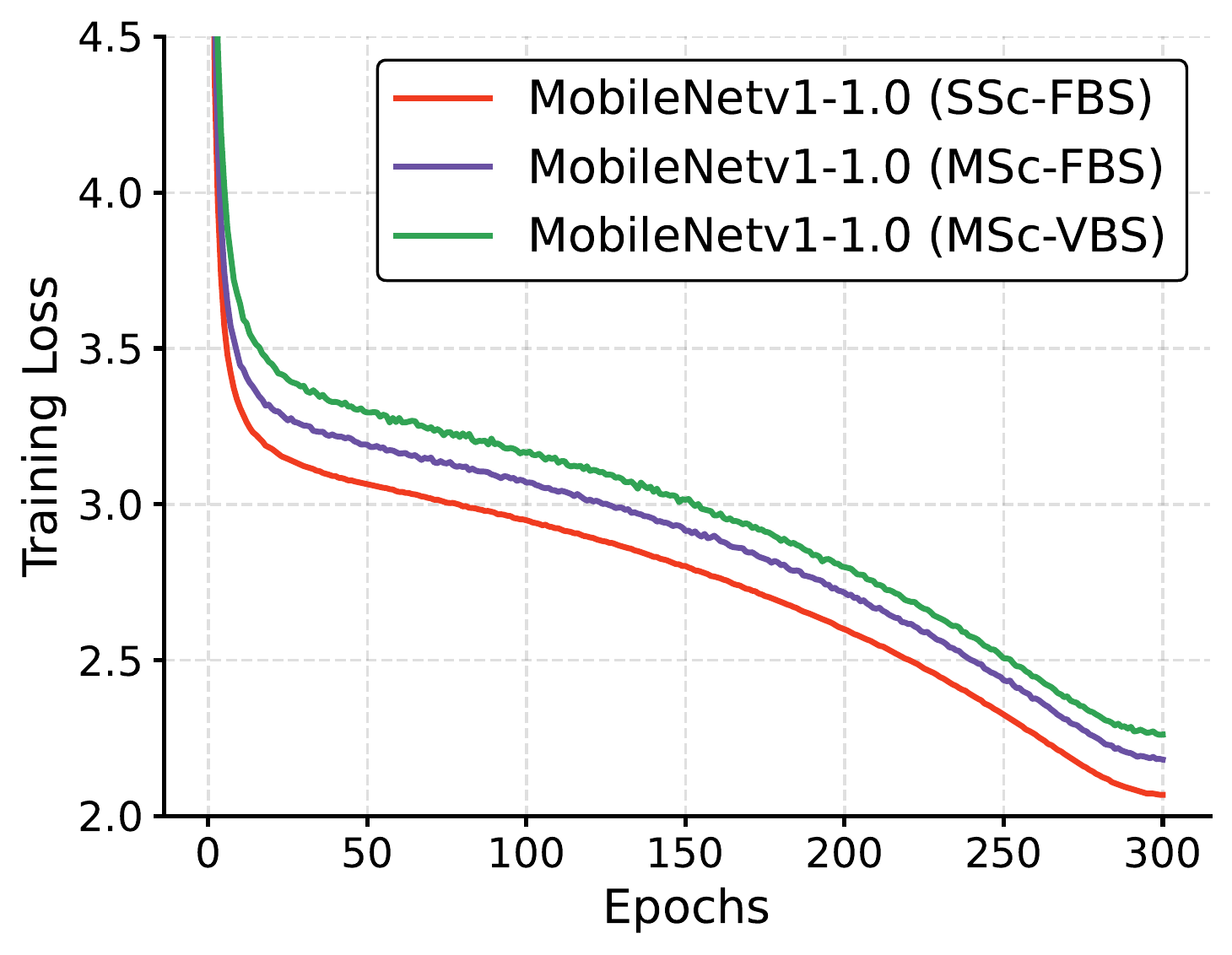} & \includegraphics[height=80px]{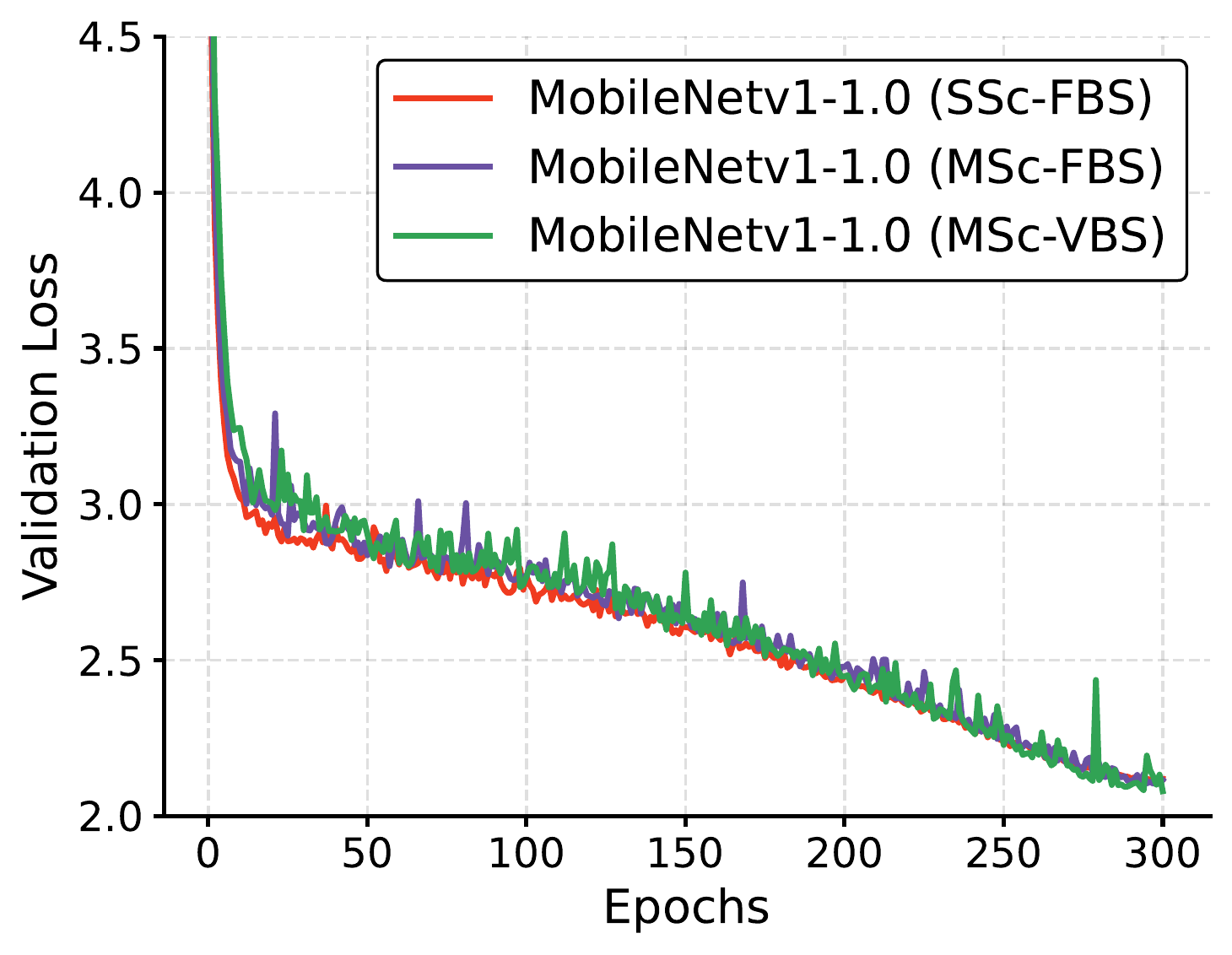} && \includegraphics[height=80px]{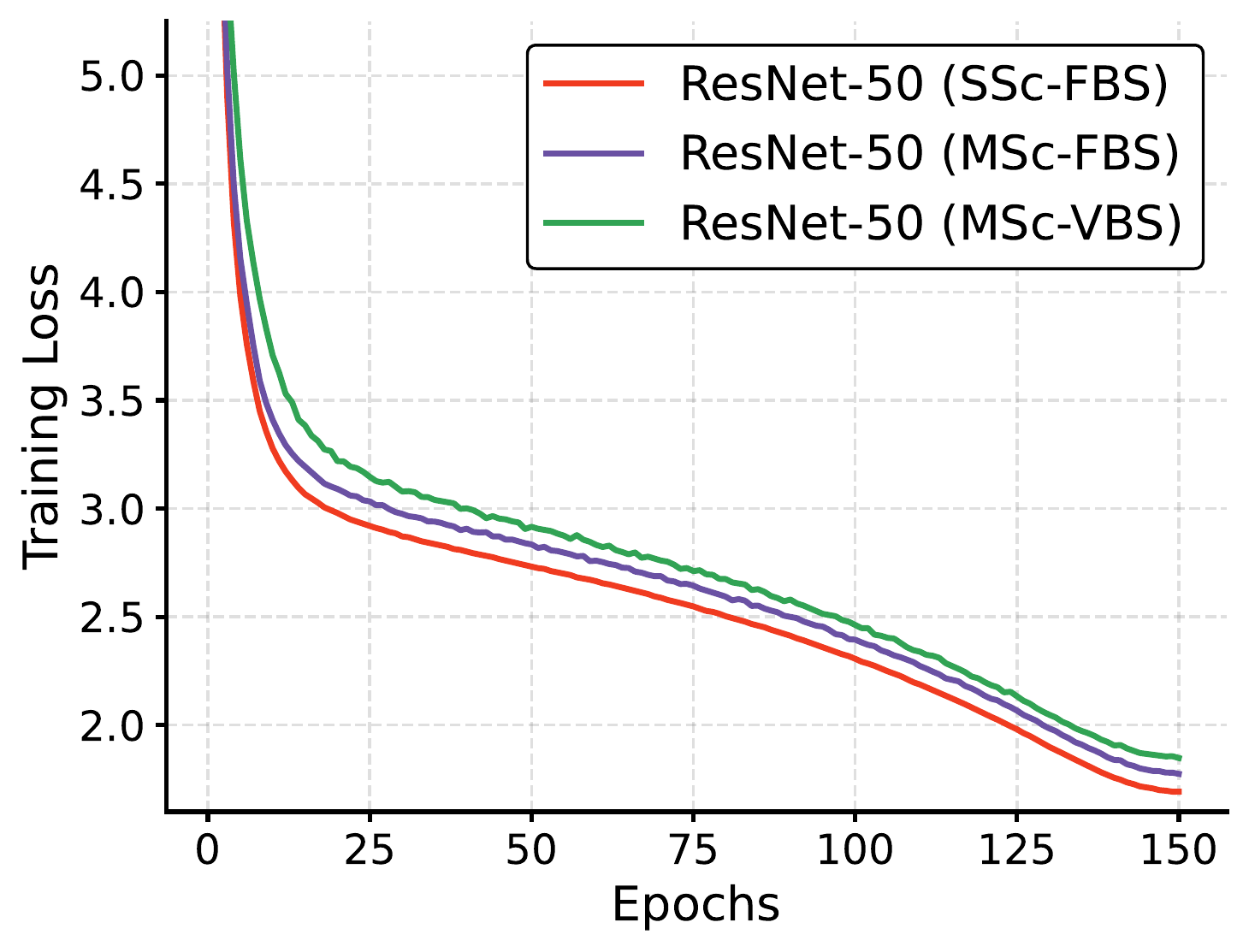} & \includegraphics[height=80px]{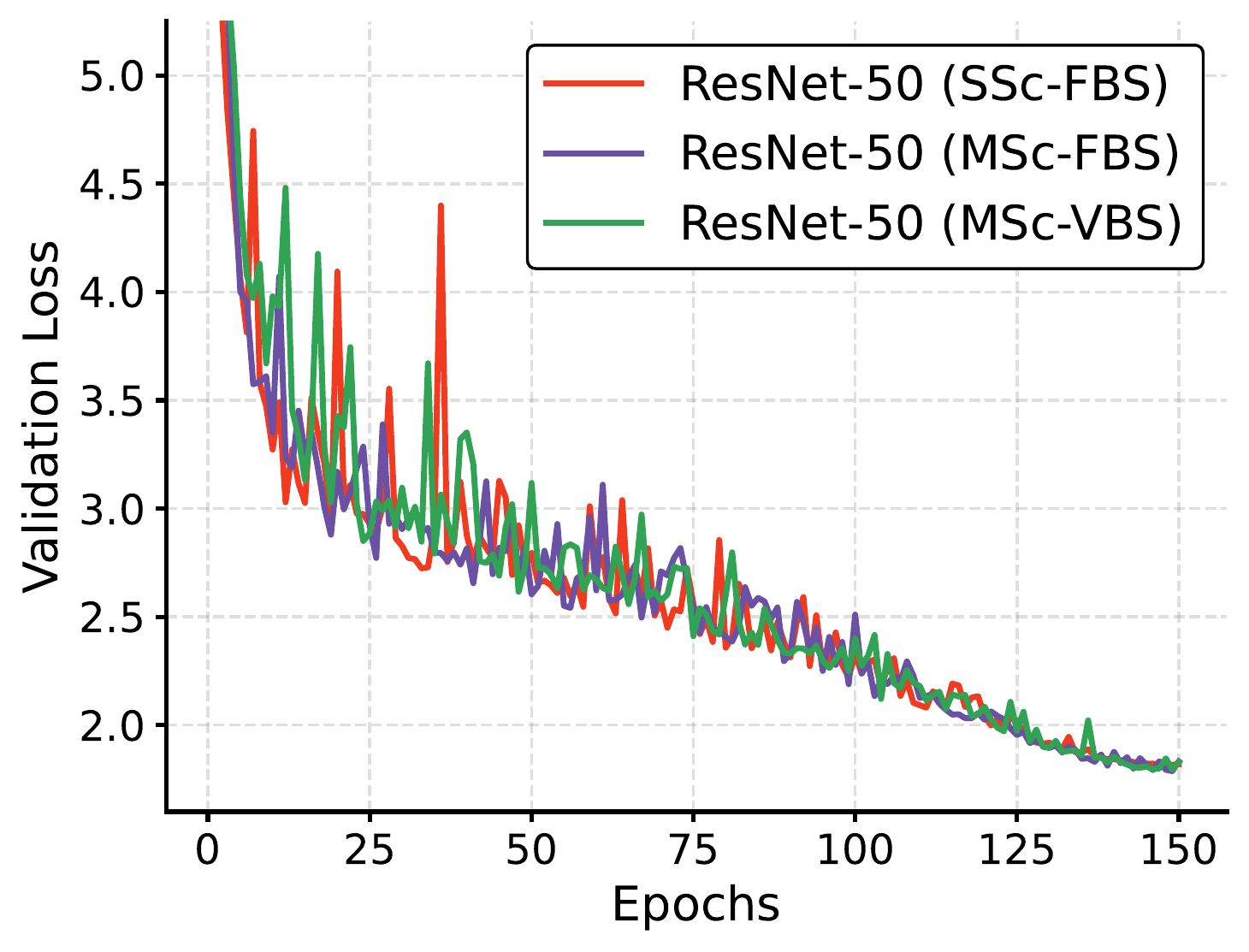} \\
             \multicolumn{2}{c}{MobileNetv1-1.0} && \multicolumn{2}{c}{ResNet-50}
        \end{tabular}
        \caption{Training and validation loss curves for models trained with different samplers}
        \label{fig:effect_sampler_train_val_loss}
    \end{subfigure}
    \vfill
    \begin{subfigure}[b]{1.2\columnwidth}
        \centering
        \resizebox{0.65\columnwidth}{!}{
        \begin{tabular}{lcccc}
            \toprule[1.5pt]
            \textbf{Sampler} & \textbf{\# optim.} & \textbf{Max. GPU} & \textbf{Train time} & \textbf{Top-1} \\
            \textbf{Sampler} & \textbf{updates} & \textbf{Memory} & \textbf{(in sec)} & @ \textbf{224x224 (\%)} \\
             \midrule[1.25pt]
             \multicolumn{5}{c}{\bfseries MobileNetv1-1.0} \\
             \midrule
              SSc-FBS & 751k \textsubscript{($1.00\times$)} & \textbf{15.9 GB} \textsubscript{($1.00\times$)} & 122k \textsubscript{($1.00\times$)} &  73.95 \textsubscript{(\textcolor{white}{+}0.00)} \\
             MSc-FBS & 751k \textsubscript{($1.00\times$)} & 26.0 GB \textsubscript{($1.64\times$)} & 137k \textsubscript{($1.12\times$)} &  \textbf{74.11} \textsubscript{(+0.16)} \\
             MSc-VBS & \textbf{574k} \textsubscript{($0.76\times$)} & 18.1 GB \textsubscript{($1.14\times$)}  & 87k \textsubscript{($0.71\times$)} & 74.05 \textsubscript{(+0.10)} \\
             \midrule[1.25pt]
             \multicolumn{4}{c}{\bfseries ResNet-50} \\
             \midrule
             SSc-FBS & 188k \textsubscript{($1.00\times$)} & \textbf{24.1 GB} \textsubscript{($1.00\times$)} & 43k \textsubscript{($1.00\times$)} &  78.12 \textsubscript{(\textcolor{white}{+}0.00)} \\
             MSc-FBS & 188k \textsubscript{($1.00\times$)} & 43.5 GB \textsubscript{($1.81\times$)} & 48k \textsubscript{($1.12\times$)} &  \textbf{78.81} \textsubscript{(+0.69)} \\
             MSc-VBS & \textbf{143k} \textsubscript{($0.79\times$)} & 28.0 GB \textsubscript{($1.16\times$)} & 36k \textsubscript{($0.84\times$)} & 78.44 \textsubscript{(+0.32)} \\
             \bottomrule[1.5pt]
        \end{tabular}
    }
    \caption{Effect of different samplers on the training performance of different models.}
    \label{fig:sampler_perf_cost}
    \end{subfigure}
    \hfill
    \begin{subfigure}[b]{0.78\columnwidth}
        \centering
        \includegraphics[height=80px]{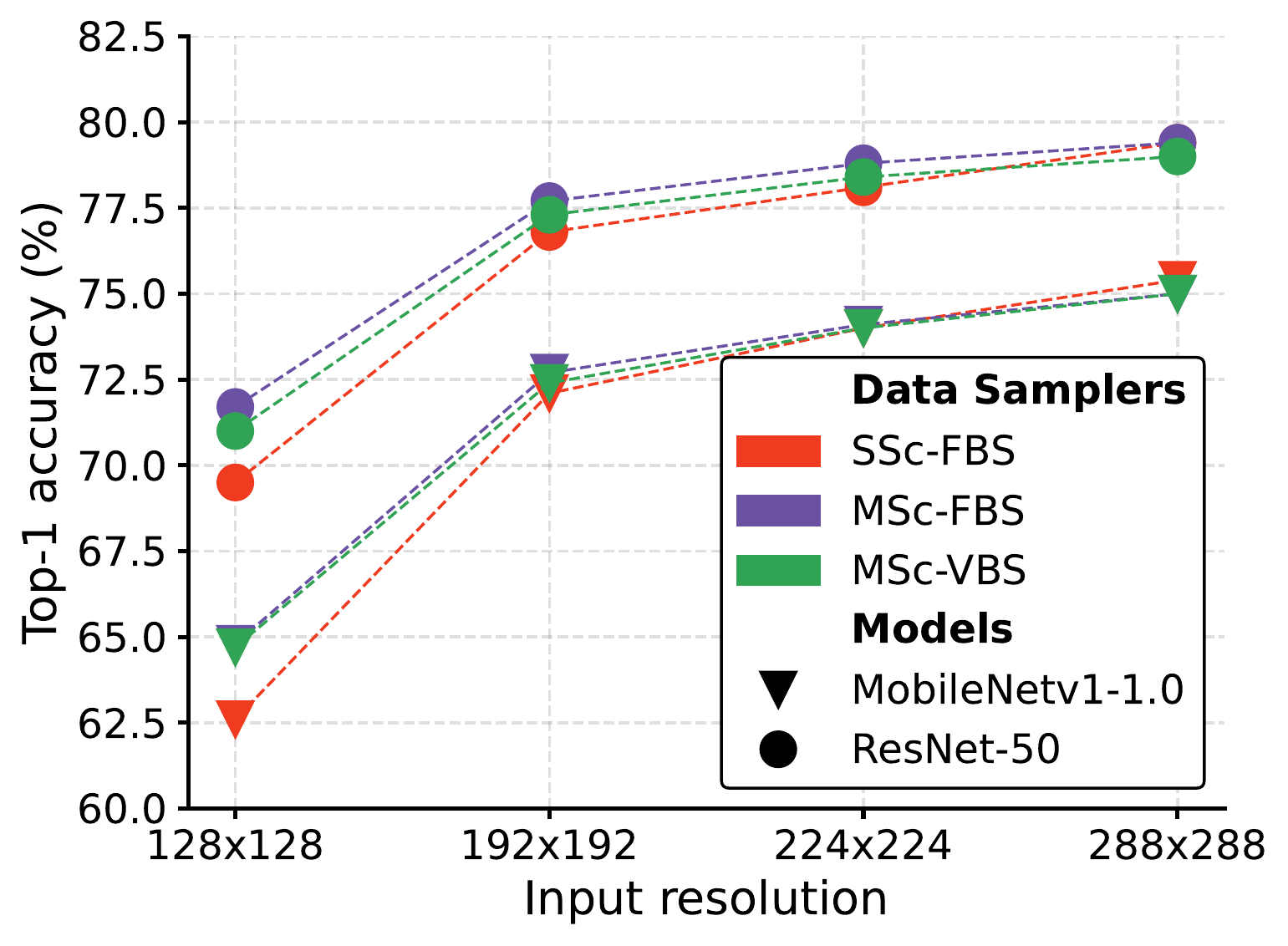}
        \caption{Effect on the performance of different models when evaluated at different input resolutions.}
    \end{subfigure}
    \caption{Effect of training deep learning models with different sampling methods on the ImageNet dataset. Models trained with MSc-VBS delivers \hlyellow{similar performance,  trains faster with fewer optimization updates, and generalizes better (higher train loss; similar validation loss)} as compared to the ones trained with SSc-FBS and MSc-FBS. Notably, models trained with MSc-VBS  require similar computational resources as SSc-FBS. Here, all models are trained with simple training recipes. For more results, see Appendix \ref{sec:app_samplers}.}
\end{figure*}

\section{\repo~Library Components}
\repo~include efficient data sampling (Section \ref{ssec:data_samplers}) and training methods (Section \ref{ssec:set}), in addition to standard components (e.g., optimizers; Section \ref{ssec:standard_comp}), which are discussed below. 

\subsection{Data Samplers}
\label{ssec:data_samplers}
\repo~offer data samplers with three sampling strategies: (1) single-scale with fixed batch size, (2) multi-scale with fixed batch size, and (3) multi-scale with variable batch size. These sampling strategies are visualized in Figure \ref{fig:data_samplers} and discussed below:

\begin{itemize}[leftmargin=*]
    \item \textbf{Single-scale with fixed batch size (SSc-FBS):} This method is the default sampling strategy in most deep learning frameworks (e.g., PyTorch, Tensorflow, and MixNet) and libraries built on top of them (e.g., the \texttt{timm} library \cite{wightman2021resnet}). At the $t$-th training iteration, this method samples a batch of $b$ images per GPU with a pre-defined spatial resolution of height $H$ and width $W$. 
    \item \textbf{Multi-scale with fixed batch size (MSc-FBS):} \sloppy The SSc-FBS method allows a network to learn representations at a single scale (or resolution). However, objects in the real-world are composed at different scales. To allow a network to learn representations at multiple scales, MSc-FBS extends SSc-FBS to multiple scales \cite{redmon2017yolo9000}. Unlike the SSc-FBS method that takes a pre-defined spatial resolution as an input, this method takes a sorted set of $n$ spatial resolutions $\mathcal{S} = \{ (H_1, W_1), (H_2, W_2), \cdots, (H_n, W_n)\}$ as an input. At the $t$-th iteration, this method randomly samples $b$ images per GPU of spatial resolution $(H_t, W_t) \in \mathcal{S}$.
    \item \textbf{Multi-scale with variable batch size (MSc-VBS):} Networks trained using the MSc-FBS methods are more robust to scale changes as compared to SSc-FBS \cite{mehta2022mobilevit}. However, depending on the maximum spatial resolution in $\mathcal{S}$, MSc-FBS methods may have a higher peak GPU memory utilization (see Figure \ref{fig:sampler_perf_cost}) as compared to SSc-FBS; causing out-of-memory errors on GPUs with limited memory. For example, MSc-FBS with $\mathcal{S} = \{ (128, 128), (192, 192), (224, 224), (320, 320)\}$ and $b=256$ would need about $2\times$ more GPU memory (for images only) than SSc-FBS with a spatial resolution of $(224, 224)$ and $b=256$. To address this memory issue, we extend MSc-FBS to variably-batch sizes in our previous work \cite{mehta2022mobilevit}. For a given sorted set of spatial resolutions $\mathcal{S} = \{ (H_1, W_1), (H_2, W_2), \cdots, (H_n, W_n)\}$ and a batch size $b$ for a maximum spatial resolution of $(H_n, W_n)$, a spatial resolution $(H_t, W_t) \in \mathcal{S}$ with a batch size of $b_t = H_n W_n b / H_t W_t$ is sampled randomly at $t$-th training iteration on each GPU.
\end{itemize}

These samplers offer different training costs and performance for different models, as shown in Figure \ref{fig:sampler_perf_cost}. Compared to SSc-FBS and MSc-FBS, MSc-VBS is a memory-efficient sampler that speeds-up the training significantly while maintaining performance.

\paragraph{\bfseries Variably-sized video sampler} Samplers discussed above can be easily extended for videos. \repo~provide variably-sized sampler for videos, wherein different video-related input variables (e.g., number of frames, number of clips per video, and spatial size) can be controlled for learning space- and time-invariant representations.

\subsection{Sample efficient training}
\label{ssec:set}
Previous works \cite{pmlr-v139-killamsetty21a-gradmatch,Killamsetty_GLISTER_2021,pmlr-v119-mirzasoleiman20a-craig} remove and re-weight data samples to reduce optimization updates (or number of forward passes) at the cost of performance degradation. Moreover, these methods are compute- and memory-intensive, and do not scale well to large models and datasets \cite{pmlr-v139-killamsetty21a-gradmatch}. This work aims to reduce optimization updates with minimal or no performance degradation.

Models learn quickly during the initial phase of training (see Figure \ref{fig:effect_sampler_train_val_loss}). In other words, models can accurately classify many training data samples during earlier epochs and such samples do not contribute much to learning. Therefore, a natural question arises: \emph{Can we remove such samples to reduce total optimization updates?}  

\begin{figure*}[t!]
    \centering
    \begin{subfigure}[b]{0.68\columnwidth}
        \centering
        \includegraphics[height=100px]{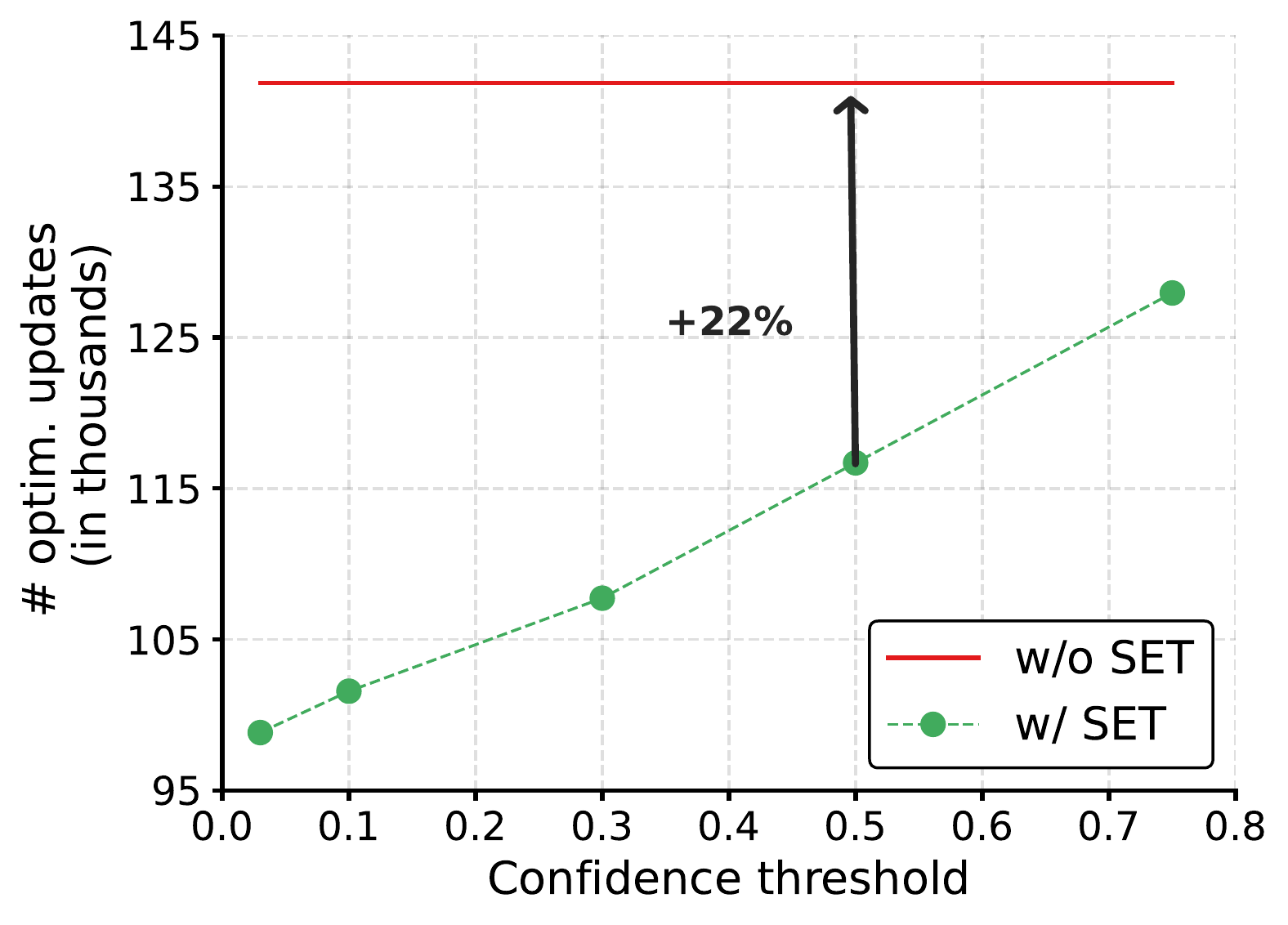}
        \caption{Optim. updates vs. confidence threshold $\tau$}
        \label{fig:set_confidence_updates}
    \end{subfigure}
    \hfill
    \begin{subfigure}[b]{0.68\columnwidth}
        \centering
        \includegraphics[height=100px]{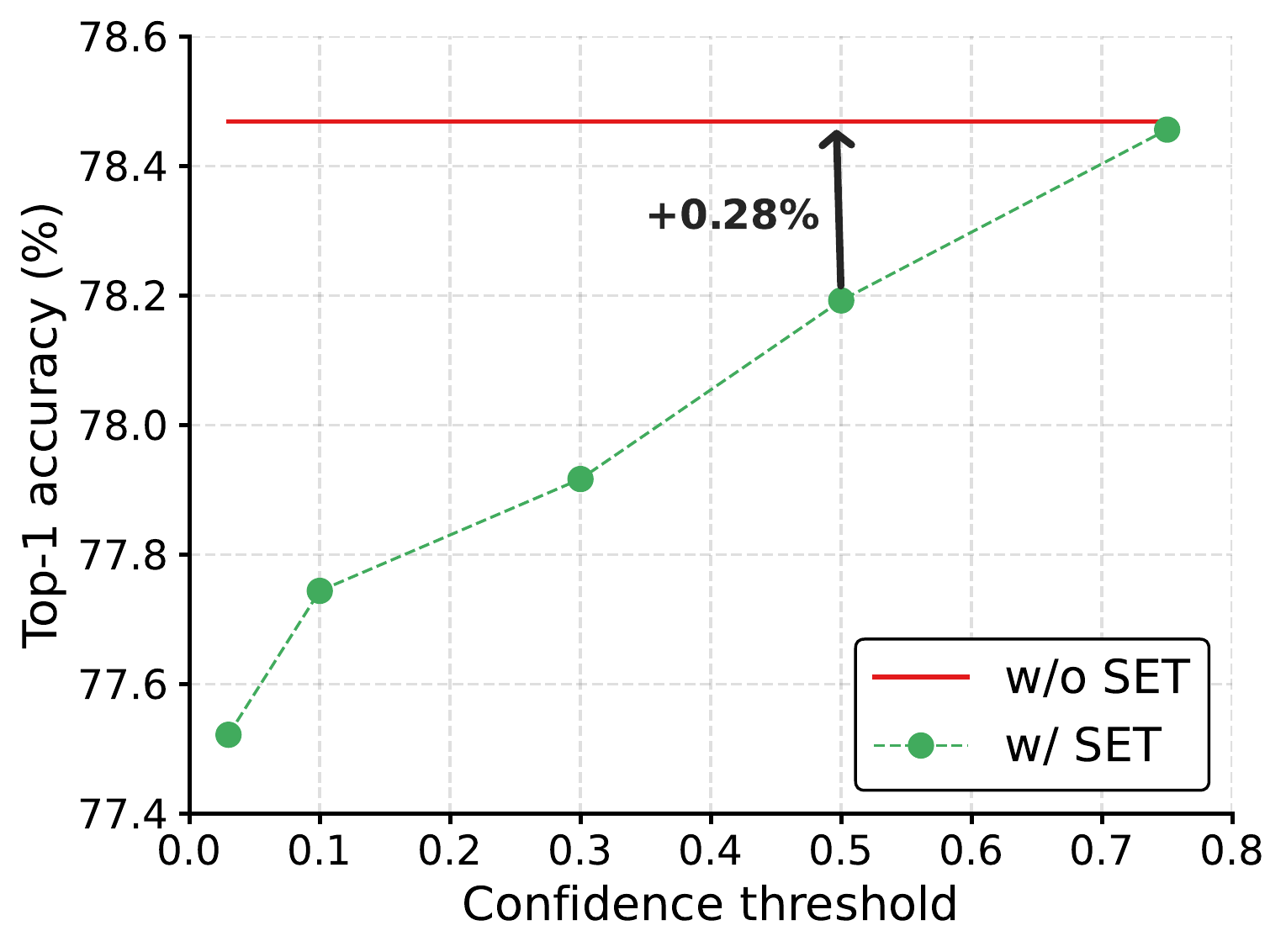}
        \caption{Top-1 accuracy vs. confidence threshold $\tau$}
        \label{fig:set_confidence_top1}
    \end{subfigure}
    \hfill
    \begin{subfigure}[b]{0.65\columnwidth}
        \centering
        \includegraphics[height=100px]{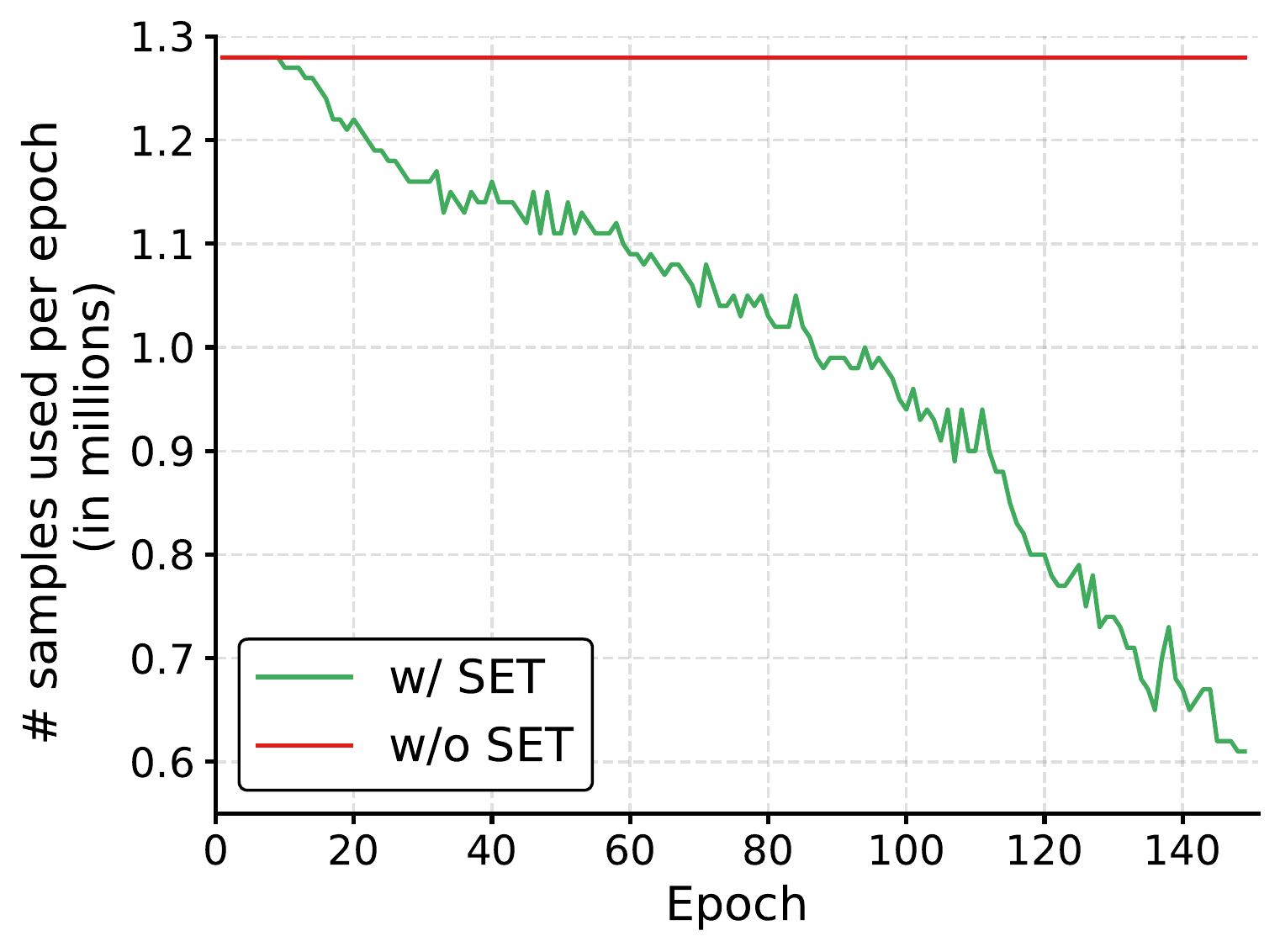}
        \caption{Training samples vs. epoch at $\tau=0.5$}
        \label{fig:set_num_samples}
    \end{subfigure}
    \caption{Sample efficient training for the ResNet-50 model. SET reduces the optimizer updates (a) while maintaining performance (b). Fluctuations in (c) represents that easy samples were classified as hard and added back to training data.}
    \label{fig:set_figure}
\end{figure*}

\repo~answer this question with a simple heuristic method, which we call \emph{Sample Efficient Training (SET)}. At each epoch, SET categorizes each sample as either \emph{hard} or \emph{easy}. To do so, SET uses a simple heuristic: if model predicts the training data sample correctly with a confidence greater than a pre-defined threshold $\tau$, then it is an easy sample and we remove it from the training data. At each epoch, model only trains using hard samples. Because of randomness in training  due to data augmentation (e.g., random cropping), it is possible that region of interest corresponding to the object category may be partially (or not) present in model's input and model may classify such inputs correctly. To make SET robust to such randomness, a training sample is classified easy only when it is classified correctly with a confidence greater than $\tau$ for a moving window of $w$ epochs. Because SET uses a window of $w$ epochs to determine easy samples, it is possible that some samples classified easy earlier may be classified harder during the later training stages. SET adds such samples back to the training data (see Figure \ref{fig:set_num_samples}). 

Figure \ref{fig:set_figure} shows results for ResNet-50 trained with and without SET using MSc-VBS. ResNet-50 without SET requires 22\% more optimization updates while delivering similar performance; demonstrating the effectiveness of SET on top of MSc-VBS. Note that SET has an overhead. Therefore, the reduction in optimization updates do not translate to reduction in training time. We believe SET can serve as a baseline in this direction and inspire future research to improve training speed while maintaining performance. 

\subsection{Standard components}
\label{ssec:standard_comp}
\repo~support different tasks (e.g., image classification, detection, segmentation), data augmentation methods (e.g., flipping, random resized crop, RandAug, and CutMix), datasets (e.g., ImageNet-1k/21k for image classification, Kinetics-400 for video classification, MS-COCO for object detection, and ADE20k for segmentation), optimizers (e.g., SGD, Adam, and AdamW), and learning rate annealing methods (e.g., fixed, cosine, and polynomial). 

\section{Benchmarks}
\label{sec:benchmarks}
\repo~support different visual recognition tasks, including  classification, detection, and segmentation. We provide comprehensive benchmarks for standard methods along with pre-trained weights.

\paragraph{\bfseries Classification on ImageNet dataset} \repo~implement popular light- and heavy-weight image classification models. The performance of some of these models on the ImageNet dataset is shown in Table \ref{tab:imagenet_1k}. With \repo, we are able to achieve better performance (e.g., MobileNetv1/v2) or similar performance (ResNet-50/101) with fewer optimization updates (faster training). See Appendix \ref{sec:app_training_procedure}, including training recipe comparisons.

\begin{table}[t!]
    \centering
    \resizebox{0.9\columnwidth}{!}{
        \begin{tabular}{lrrcl}
            \toprule[1.5pt]
            \multirow{2}{*}{\bfseries Model} & \multirow{2}{*}{\bfseries Params} & \multirow{2}{*}{\bfseries FLOPs} & \multicolumn{2}{c}{\bfseries Top-1 (in \%)} \\
            \cmidrule[1.5pt]{4-5}
             & & & {\bfseries \repo~(Ours)} & {\bfseries Prev.} \\
            \midrule[1.25pt]
            MobileNetv1-0.25 & 0.5 M & 46.3 M & \textbf{54.5} & 49.8 \\
            MobileNetv1-1.0 & 4.2 M & 568.7 M & \textbf{74.1} & 70.9 \\
            \midrule
            MobileNetv2-0.25 & 1.5 M & 50.5 M & \textbf{53.6} & -- \\
            MobileNetv2-1.0 & 3.5 M & 300.7 M & \textbf{72.9} & 72.0 \\
            \midrule
            \multirow{2}{*}{MobileNetv3-Large} & \multirow{2}{*}{5.4 M} & \multirow{2}{*}{210.7 M} & \multirow{2}{*}{75.1} & \textbf{75.2} \textsubscript{(TPU)} \\
             &  &  & & 74.6 \textsubscript{(GPU)} \\
            \midrule
            ResNet-101 (simple recipe) & 44.5 M & 7.7 G & 79.8 & 77.4$^\dagger$ \\
            ResNet-101 (adv. recipe) & 44.5 M & 7.7 G & 81.8 & 81.8$^\dagger$ \\
            \midrule
            ViT-Tiny & 5.7 M & 1.3 G & 72.9 & 72.2$^\star$ \\
            \bottomrule[1.5pt]
        \end{tabular}
    }
    \caption{Classification on the ImageNet dataset. \\$^\dagger$\small{Torchvision \cite{pytorch2022pytorch} requires $1.3\times$ more optimization updates (forward passes) as compared to \repo; see Appendix \ref{sec:app_training_procedure} for details.} $^\star$ Results are from \cite{touvron2021training}.}
    \label{tab:imagenet_1k}
\end{table}

\paragraph{\bfseries Detection and segmentation} Similar to image classification, \repo~can be used to train standard detection and segmentation models with better performance. For example, SSD with ResNet-101 backbone trained with \repo~at a resolution of $384\times 384$ delivers a 1.6\% better mAP than the same model trained at a resolution of $512 \times 512$ as reported in \cite{fu2017dssd}. Similarly, on the task of semantic segmentation on the ADE20k dataset using DeepLabv3 with MobileNetv2 as the backbone, \repo~delivers 1.1\% better performance than \texttt{MMSegmentation} library \cite{mmseg2020} with $2\times$ fewer epochs and optimization updates. For more details, please see our benchmarking results at \url{https://github.com/apple/ml-cvnets}.

\section{Conclusion}
This paper introduces \repo, a modular deep learning library for visual recognition tasks with high performance. In future, we plan to continue enhancing \repo~with novel and reproducible methods. We welcome contributions from the research and open-source community to support further innovation.


\bibliographystyle{ACM-Reference-Format}
\bibliography{main}

\clearpage

\appendix


\section{Comparison of training procedures}
\label{sec:app_training_procedure}

Training ResNet-50 \cite{he2016deep} on the ImageNet dataset \cite{russakovsky2015imagenet} is widely used for comparing training procedures \cite{pytorch2022pytorch, wightman2021resnet}. Table \ref{tab:resnet_training_procedure_comparison} compares the performance of \repo' training recipe with previous methods. We can make following observations:

\begin{table*}[b!]
    \centering
    \resizebox{2.1\columnwidth}{!}{
        \begin{tabular}{l|ccccccc|ccc|c}
            \toprule[1.5pt]
             &  \textbf{ResNet} \cite{he2016deep} & \textbf{Torchvision} \cite{pytorch2022pytorch} & \textbf{FixRes} \cite{touvron2019fixing} & \textbf{DeiT} \cite{touvron2021training} & \textbf{timm-A3} \cite{wightman2021resnet} & \textbf{timm-A1} \cite{wightman2021resnet}  & \textbf{Torchvision-Adv} \cite{pytorch2022pytorch} & \multicolumn{4}{c}{\textbf{\repo~(Ours)}} \\
             \midrule[1pt]
             Train res & 224 & 224 & 224 & 224 & 160 & 224 & 176 & 224 & \multicolumn{3}{c}{\{128, 192, 224, 288, 320\}} \\
             Test res & 224 & 224 & 224 & 224 &  224 & 224 & 224 & 224 & 224 & 224 & 224 \\
             Test crop ratio & 0.875 & 0.875 &  0.875 & 0.875 & 0.95 & 0.95 & 0.95 & 0.875 & 0.875 & 0.875 & 0.95 \\
             Data sampler & SSc-FBS & SSc-FBS & SSc-FBS & SSc-FBS & SSc-FBS &  SSc-FBS & SSc-FBS & SSc-FBS & MSc-FBS & MSc-VBS & MSc-VBS \\
             \midrule
             Epochs & 90 & 90 & 120 & 300 & 100 & 600 & \cellcolor{red!20}{600} & 150 & 150 & 150 & \cellcolor{red!20}{600} \\
             Batch size & 256 & 256 & 512 & 1024 & 2048 & 2048 & \cellcolor{red!20}{1024} & 1024 & 1024 & 1024 & \cellcolor{red!20}{1024} \\
             \# optim. updates & 450k & 450k & 300k & 375k & 63k & 375k & \cellcolor{red!20}{750k} & 188k & 188k & 143k & \cellcolor{red!20}{571k} \\
             \midrule
             Max. LR & 0.1 & 0.1 & 0.2 & 0.001 & 0.008 & 0.005 & 0.5 & 0.4 & 0.4 & 0.4 & 0.4 \\
             LR Annealing & step & step & step & cosine & cosine & cosine & cosine & cosine & cosine & cosine & cosine \\
             Weight decay & $10^{-4}$ & $10^{-4}$ & $10^{-4}$ & 0.05 & 0.02 & 0.01 & $2^{-5}$ & $10^{-4}$ & $10^{-4}$ & $10^{-4}$ & $10^{-4}$\\
             Warmup epochs & 0 & 0 & 0 & 5 & 5 & 5 & 5 & 5 & 5 & 5 & 5 \\
             Optimizer & SGD-M & SGD-M & SGD-M & AdamW & LAMB & LAMB & SGD-M & SGD-M & SGD-M & SGD-M & SGD-M \\
             Loss fn. & CE & CE & CE & CE & BCE & BCE & CE & CE & CE & CE & CE\\
             EMA & \xmark & \xmark & \xmark & \xmark & \xmark & \xmark & \cmark & \cmark & \cmark & \cmark & \cmark\\
             \midrule
             Label smoothing $\epsilon$ & \xmark & \xmark & \xmark & 0.1 & \xmark &  0.1 & 0.1 & 0.1 & 0.1 & 0.1 & 0.1 \\
             Stoch. Depth & \xmark & \xmark & \xmark & 0.1 & \xmark & \cmark & \xmark & \xmark & \xmark & \xmark & \xmark \\
             Mixed Precision & \xmark & \xmark & \xmark & \cmark & \cmark & \cmark & \cmark & \cmark & \cmark & \cmark & \cmark \\
             \midrule
             H. flip & \cmark & \cmark & \cmark & \cmark & \cmark & \cmark & \cmark  & \cmark & \cmark & \cmark & \cmark \\
             RRC & \xmark & \cmark & \cmark & \cmark & \cmark & \cmark  & \cmark  & \cmark & \cmark  & \cmark  & \cmark\\
             Repeated Aug \cite{hoffer2019augment} & \xmark & \xmark & \cmark &  \cmark & \xmark & \cmark & \xmark & \xmark & \xmark & \xmark & \xmark\\
             Rand Augment \cite{cubuk2020randaugment} & \xmark & \xmark & \xmark & \cmark & \cmark & \cmark & \cmark & \xmark & \xmark & \xmark & \cmark\\
             Mixup \cite{zhang2017mixup} & \xmark & \xmark & \xmark & \cmark & \cmark & \cmark & \cmark & \xmark & \xmark & \xmark & \cmark \\
             Cutmix \cite{yun2019cutmix} & \xmark & \xmark & \xmark & \cmark & \cmark & \cmark & \cmark & \xmark & \xmark & \xmark & \cmark\\
             Random Erase \cite{zhong2020random} & \xmark & \xmark & \xmark & \cmark  & \xmark & \xmark & \cmark & \xmark & \xmark & \xmark & \cmark \\
             ColorJitter & \xmark & \cmark & \cmark & \xmark  & \xmark & \xmark & \xmark & \xmark & \xmark & \xmark & \xmark\\
             PCA lighting & \cmark & \xmark & \xmark & \xmark & \xmark & \xmark & \xmark & \xmark & \xmark & \xmark & \xmark\\
             \midrule
             Top-1 acc. & 75.3\% & 76.1\% & 77.0\% & 78.4\% & 78.1\% & \textbf{80.4\%} & \textbf{80.4\%} & 78.1\% & 78.8\% & 78.4\% & \textbf{80.4\%} \\
            \bottomrule[1.5pt]
        \end{tabular}
    }
    \caption{Comparison of ResNet-50 training recipes and performance in different papers with \repo (ours). \\
    \small{Link to Torchvision-Adv recipe: \url{https://pytorch.org/blog/how-to-train-state-of-the-art-models-using-torchvision-latest-primitives/}}. Last accessed on May 9, 2022.}
    \label{tab:resnet_training_procedure_comparison}
\end{table*}

\begin{itemize}[leftmargin=*]
    \item \textbf{Simple training recipes} are useful for research in resource-constrained environments, where number of GPUs available for training are limited. With simple recipes, ResNet-50 trained with \repo~achieves better performance as compared to previous works (e.g., ResNet, Torchvision, and FixRes in Table \ref{tab:resnet_training_procedure_comparison}).
    \item \textbf{Advanced training recipes} are useful for achieving state-of-the-art performance, as they train models for longer with better augmentation methods. With advanced recipes, ResNet-50 trained with \repo~achieve similar performance to previous works, but with $1.3 \times$ fewer optimization updates (Torchvision-Adv vs. \repo' recipe; see highlighted text in Table \ref{tab:resnet_training_procedure_comparison}). 
\end{itemize}

\paragraph{\bfseries Comparison with light-weight model training recipes} Table \ref{tab:mobilenetv2_training_procedure_comparison} and Table \ref{tab:mobilenetv3_training_procedure_comparison} compares the performance of MobileNetv2 and MobileNetv3 models trained with different training recipes. With simple training recipes, models trained with \repo achieve similar or better performance as compared to other libraries (e.g., TensorflowLite or Torchvision). For example, both \repo and Torchvision are able to train MobileNetv3 with reported performance of 75.2\% \cite{howard2019searching}. However, \repo~achieve the reported performance with basic data augmentation while Torchvision achieves this with more data augmentation (e.g., RandAugment, Cutmix, MixUp, and RandomErase).

\begin{table*}[t!]
    \centering
    \resizebox{1.6\columnwidth}{!}{
        \begin{tabular}{l|cc|ccc}
            \toprule[1.5pt]
             &  \textbf{Torchvision} \cite{pytorch2022pytorch} & \textbf{TensorflowLite} \cite{sandler2018mobilenetv2} & \multicolumn{3}{c}{\textbf{\repo~(Ours)}} \\
             \midrule[1pt]
             Train res & 224 & 224 & 224  & \{128, 192, 224, 288, 320\} & \{128, 192, 224, 288, 320\} \\
             Test res & 224 & 224 & 224 & 224 & 224 \\
             Test crop ratio & 0.875 & 0.875  & 0.875 & 0.875 & 0.875 \\
             Data sampler & SSc-FBS & SSc-FBS & SSc-FBS & MSc-FBS & MSc-VBS \\
             \midrule
             Epochs & \cellcolor{red!20}{300} & \cellcolor{red!20}{450} & 300 & 300 & \cellcolor{red!20}{300} \\
             Batch size & \cellcolor{red!20}{256} & \cellcolor{red!20}{768} & 1024 & 1024 & \cellcolor{red!20}{1024} \\
             \# optim. updates & \cellcolor{red!20}{1500k} & \cellcolor{red!20}{700k} & 375k & 375k & \cellcolor{red!20}{288k} \\
             \midrule
             Max. LR & 0.36 & 0.045 & 0.4 & 0.4 & 0.4 \\
             LR Annealing & poly & poly & cosine & cosine & cosine \\
             Weight decay & $4^{-5}$ & $4^{-5}$ & $4^{-5}$ & $4^{-5}$ & $4^{-5}$ \\
             Warmup epochs & 0 & 0 & 5 & 5 & 5  \\
             Optimizer & SGD-M & RMSProp & SGD-M & SGD-M & SGD-M \\
             Loss fn. & CE & CE & CE & CE & CE\\
             EMA & \xmark & \cmark & \cmark & \cmark & \cmark\\
             Label smoothing $\epsilon$ & 0.0 & 0.1 & 0.1 & 0.1 & 0.1 \\
             \midrule
             H. flip & \cmark & \cmark & \cmark & \cmark  & \cmark \\
             RRC & \cmark & \cmark & \cmark  & \cmark  & \cmark\\
             \midrule
             Top-1 acc. & 71.8\% & 72.0\% & 73.3\% & \textbf{73.4}\% & 72.9\% \\
            \bottomrule[1.5pt]
        \end{tabular}
    }
    \caption{Comparison of MobileNetv2-1.0 training recipes and performance in different libraries with \repo (ours).}
    \label{tab:mobilenetv2_training_procedure_comparison}
\end{table*}

\begin{table*}[t!]
    \centering
    \resizebox{1.6\columnwidth}{!}{
        \begin{tabular}{l|cc|ccc}
            \toprule[1.5pt]
             &  \textbf{Torchvision-adv} \cite{pytorch2022pytorch} & \textbf{TensorflowLite} \cite{howard2019searching} & \multicolumn{3}{c}{\textbf{\repo~(Ours)}} \\
             \midrule[1pt]
             Train res & 224 & 224 & 224  & \{128, 192, 224, 288, 320\} & \{128, 192, 224, 288, 320\} \\
             Test res & 224 & 224 & 224 & 224 & 224 \\
             Test crop ratio & 0.875 & 0.875  & 0.875 & 0.875 & 0.875 \\
             Data sampler & SSc-FBS & SSc-FBS & SSc-FBS & MSc-FBS & MSc-VBS \\
             \midrule
             Epochs & \cellcolor{red!20}{600} & \cellcolor{red!20}{840$^\dagger$} & 300 & 300 & \cellcolor{red!20}{300} \\
             Batch size & \cellcolor{red!20}{1024} & \cellcolor{red!20}{1536} & 2048 & 2048 & \cellcolor{red!20}{2048} \\
             \# optim. updates & \cellcolor{red!20}{750k} & \cellcolor{red!20}{700k} & 188k & 188k & \cellcolor{red!20}{144k} \\
             \midrule
             Max. LR & 0.064 & 0.16 & 0.4 & 0.4 & 0.4 \\
             LR Annealing & poly & poly & cosine & cosine & cosine \\
             Weight decay & $1^{-5}$ & $1^{-5}$ & $4^{-5}$ & $4^{-5}$ & $4^{-5}$ \\
             Warmup epochs & 0 & 5 & 5 & 5 & 5  \\
             Optimizer & RMSProp & RMSProp & SGD-M & SGD-M & SGD-M \\
             Loss fn. & CE & CE & CE & CE & CE\\
             EMA & \xmark & \cmark & \cmark & \cmark & \cmark\\
             Label smoothing $\epsilon$ & 0.0 & 0.1 & 0.1 & 0.1 & 0.1 \\
             \midrule
             H. flip & \cmark & \cmark & \cmark & \cmark  & \cmark \\
             RRC & \cmark & \cmark & \cmark  & \cmark  & \cmark\\
             Rand Augment & \cmark & \xmark & \xmark & \xmark & \xmark\\
             Random Erase & \cmark & \xmark & \xmark & \xmark & \xmark\\
             Cutmix & \cmark & \xmark & \xmark & \xmark & \xmark\\
             MixUp & \cmark & \xmark & \xmark & \xmark & \xmark\\
             \midrule
             Top-1 acc. & 75.3\% & 74.6\% & 75.3\% & \textbf{75.3}\% & 75.1\% \\
            \bottomrule[1.5pt]
        \end{tabular}
    }
    \caption{Comparison of MobileNetv3-Large training recipes and performance in different libraries with \repo (ours). $^\dagger$ Epochs are computed based on the training recipe in the official repository. \\ \small{Link to official training recipe: \url{https://github.com/tensorflow/models/tree/master/research/slim/nets/mobilenet}}. Last accessed on May 9, 2022.}
    \label{tab:mobilenetv3_training_procedure_comparison}
\end{table*}

\section{Effect of different samplers}
\label{sec:app_samplers}
\repo~implement three different samplers: (1) single-scale with fixed batch size (SSc-FBS), (2) multi-scale with fixed batch size (MSc-FBS), and (3) multi-scale with variable batch size (MSc-VBS). For different models, these samplers may offer different training costs and performance. Table \ref{tab:app_effect_samplers_diff_models} compares the effect of different samplers on training efficiency and performance on ImageNet. Clearly, MSc-VBS is a memory-efficient sampler that speed-up the training significantly while delivering similar performance to other two samplers.

\begin{table*}[t!]
\resizebox{2\columnwidth}{!}{
\begin{tabular}{lccccccc}
\toprule[1.5pt]
\textbf{Model} & \multicolumn{3}{c}{\textbf{Sampler}} & \textbf{Peak GPU Memory} & \textbf{Train time} & \textbf{\# optim. updates} & \textbf{Top-1}  \\
\cmidrule[1pt]{2-4}
 & \textbf{SSc-FBS} & \textbf{MSc-FBS} & \textbf{MSc-VBS} & \textbf{(in GB)} & \textbf{(in sec)} & \textbf{(in thousands)} & \textbf{(in \%)} \\
\midrule[1.5pt]
\multirow{3}{*}{\bfseries ResNet-34} & \cmark & \xmark & \xmark & 15.7 \textsubscript{($1.00 \times$)} & 35k \textsubscript{($1.00 \times$)} & 188k \textsubscript{($1.00 \times$)} & 75.1 \textsubscript{(\textcolor{white}{+}0.0)} \\
  & \xmark  & \cmark & \xmark & 24.2 \textsubscript{($1.54 \times$)} & 40k \textsubscript{($1.14 \times$)} & 188k \textsubscript{($1.00 \times$)} & 75.2 \textsubscript{(+0.1)}  \\
    & \xmark & \xmark & \cmark & 17.8 \textsubscript{($1.13 \times$)} & 31k \textsubscript{($0.89 \times$)} & 143k \textsubscript{($0.76 \times$)} & 74.8 \textsubscript{(-0.3)}  \\
\midrule
\multirow{3}{*}{\bfseries ResNet-50} & \cmark & \xmark & \xmark & 24.1 \textsubscript{($1.00 \times$)} & 43k \textsubscript{($1.00 \times$)}  & 188k \textsubscript{($1.00 \times$)} & 78.1 \textsubscript{(\textcolor{white}{+}0.0)} \\
  & \xmark  & \cmark & \xmark & 43.5 \textsubscript{($1.80 \times$)} & 48k \textsubscript{($1.12 \times$)} & 188k \textsubscript{($1.00 \times$)} & 78.8 \textsubscript{(+0.7)} \\
    & \xmark & \xmark & \cmark & 28.0 \textsubscript{($1.16 \times$)} & 36k \textsubscript{($0.84 \times$)} & 143k \textsubscript{($0.76 \times$)} & 78.4 \textsubscript{(+0.3)} \\
\midrule
\multirow{3}{*}{\bfseries ResNet-101} & \cmark & \xmark & \xmark & 31.1 \textsubscript{($1.00 \times$)}   & 57k \textsubscript{($1.00 \times$)} & 188k \textsubscript{($1.00 \times$)} & 79.6 \textsubscript{(\textcolor{white}{+}0.0)} \\
  & \xmark  & \cmark & \xmark & 61.7 \textsubscript{($1.98 \times$)} & 64k \textsubscript{($1.12 \times$)} & 188k \textsubscript{($1.00 \times$)} & 80.0 \textsubscript{(+0.4)} \\
  & \xmark  & \xmark & \cmark  & 35.3 \textsubscript{($1.14 \times$)} & 45k \textsubscript{($0.79 \times$)} & 143k \textsubscript{($0.76 \times$)} & 79.8 \textsubscript{(+0.2)}\\
\midrule
\midrule
\multirow{3}{*}{\bfseries MobileNetv1-0.25}  & \cmark & \xmark & \xmark & 10.0 \textsubscript{($1.00 \times$)} & 122k \textsubscript{($1.00 \times$)} & 751k \textsubscript{($1.00 \times$)} & 55.5 \textsubscript{(\textcolor{white}{+}0.0)}\\
  & \xmark  & \cmark & \xmark & 13.1 \textsubscript{($1.31 \times$)} & 125k \textsubscript{($1.02 \times$)} & 751k \textsubscript{($1.00 \times$)} & 54.4 \textsubscript{(-1.1)}\\
  & \xmark  & \xmark & \cmark & 10.5 \textsubscript{($1.05 \times$)} & \textcolor{white}{1}85k \textsubscript{($0.69 \times$)} & 574k \textsubscript{($0.76 \times$)} & 54.5 \textsubscript{(-1.0)}\\
\midrule
\multirow{3}{*}{\bfseries MobileNetv1-0.50}  & \cmark & \xmark & \xmark & 12.0 \textsubscript{($1.00 \times$)} & 118k \textsubscript{($1.00 \times$)} & 751 k \textsubscript{($1.00 \times$)} & 66.6 \textsubscript{(\textcolor{white}{+}0.0)}\\
  & \xmark  & \cmark & \xmark & 16.2 \textsubscript{($1.35 \times$)} & 134k \textsubscript{($1.14 \times$)} & 751k \textsubscript{($1.00 \times$)} & 66.5 \textsubscript{(-0.1)}\\
  & \xmark  & \xmark & \cmark & 12.3 \textsubscript{($1.03 \times$)} & \textcolor{white}{1}86k \textsubscript{($0.73 \times$)} & 574k \textsubscript{($0.76 \times$)} & 65.9 \textsubscript{(-0.7)} \\
\midrule
\multirow{3}{*}{\bfseries MobileNetv1-0.75}  & \cmark & \xmark & \xmark & 13.9 \textsubscript{($1.00 \times$)} & 126k \textsubscript{($1.00 \times$)} & 751k \textsubscript{($1.00 \times$)} & 71.5 \textsubscript{(\textcolor{white}{+}0.0)} \\
  & \xmark  & \cmark & \xmark & 21.8 \textsubscript{($1.57 \times$)} & 129k \textsubscript{($1.02 \times$)} & 751k \textsubscript{($1.00 \times$)} & 71.7 \textsubscript{(+0.2)}\\
  & \xmark  & \xmark & \cmark & 15.3 \textsubscript{($1.10 \times$)}  & 114k \textsubscript{($0.90 \times$)}  & 574k \textsubscript{($0.76 \times$)} & 71.4 \textsubscript{(-0.1)}\\
\midrule
\multirow{3}{*}{\bfseries MobileNetv1-1.0} & \cmark & \xmark & \xmark & 15.9 \textsubscript{($1.00 \times$)} & 122k \textsubscript{($1.00 \times$)} & 751k \textsubscript{($1.00 \times$)} & 74.0 \textsubscript{(\textcolor{white}{+}0.0)} \\
    & \xmark & \cmark & \xmark & 26.0 \textsubscript{($1.64 \times$)} & 137k \textsubscript{($1.12 \times$)} & 751k \textsubscript{($1.00 \times$)} & 74.1 \textsubscript{(+0.1)} \\
    & \xmark & \xmark & \cmark & 18.1 \textsubscript{($1.14 \times$)} & \textcolor{white}{1}87k \textsubscript{($0.71 \times$)} & 574k \textsubscript{($0.76 \times$)} & 74.1 \textsubscript{(+0.1)}\\
\midrule
\midrule
\multirow{3}{*}{\bfseries MobileNetv2-0.25}  & \cmark & \xmark & \xmark & 21.3 \textsubscript{($1.00 \times$)}   & 119k \textsubscript{($1.00 \times$)}   & 375k \textsubscript{($1.00 \times$)} & 54.9 \textsubscript{(\textcolor{white}{+}0.0)} \\
    & \xmark & \cmark & \xmark & 33.5 \textsubscript{($1.57 \times$)} & 135k \textsubscript{($1.13 \times$)} & 375k \textsubscript{($1.00 \times$)} & 54.0 \textsubscript{(-0.9)} \\
 & \xmark & \xmark & \cmark & 25.5 \textsubscript{($1.20 \times$)}  & \textcolor{white}{1}75k  \textsubscript{($0.63 \times$)} & 288k \textsubscript{($0.77 \times$)}  & 53.6 \textsubscript{(-1.3)}\\
 \midrule
\multirow{3}{*}{\bfseries MobileNetv2-0.50}  & \cmark & \xmark & \xmark & 26.4 \textsubscript{($1.00 \times$)} & 120k \textsubscript{($1.00 \times$)} & 375k \textsubscript{($1.00 \times$)}  & 65.8 \textsubscript{(\textcolor{white}{+}0.0)} \\
 & \xmark & \cmark & \xmark & 45.2 \textsubscript{($1.71 \times$)} & 138k \textsubscript{($1.15 \times$)} & 375k \textsubscript{($1.00 \times$)} & 65.6 \textsubscript{(-0.2)}\\
 & \xmark & \xmark & \cmark & 30.9 \textsubscript{($1.17 \times$)} & \textcolor{white}{1}99k \textsubscript{($0.82 \times$)}  & 288k \textsubscript{($0.77 \times$)}  & 65.3 \textsubscript{(-0.5)}\\
 \midrule
\multirow{3}{*}{\bfseries MobileNetv2-0.75}  & \cmark & \xmark & \xmark & 34.3 \textsubscript{($1.00 \times$)} & 127k \textsubscript{($1.00 \times$)} & 375k \textsubscript{($1.00 \times$)} & 70.7 \textsubscript{(\textcolor{white}{+}0.0)} \\
 & \xmark & \cmark & \xmark & 65.3 \textsubscript{($1.90 \times$)}  & 139k \textsubscript{($1.09 \times$)} & 375k \textsubscript{($1.00 \times$)} & 70.8 \textsubscript{(+0.1)}\\
 & \xmark & \xmark & \cmark & 41.0 \textsubscript{($1.20 \times$)}  & \textcolor{white}{1}85k \textsubscript{($0.67 \times$)}  & 288k \textsubscript{($0.77 \times$)}  & 70.4 \textsubscript{(-0.3)}\\
 \midrule
\multirow{3}{*}{\bfseries MobileNetv2-1.0}   & \cmark & \xmark & \xmark & 36.8  \textsubscript{($1.00 \times$)} & 128k \textsubscript{($1.00 \times$)} & 375k \textsubscript{($1.00 \times$)} & 73.3 \textsubscript{(\textcolor{white}{+}0.0)}\\
    & \xmark & \cmark & \xmark & 70.7 \textsubscript{($1.92 \times$)} & 142k \textsubscript{($1.11 \times$)} & 375k \textsubscript{($1.00 \times$)} & 73.4 \textsubscript{(+0.1)}\\
   & \xmark & \xmark & \cmark & 43.9 \textsubscript{($1.19 \times$)} & \textcolor{white}{1}86k \textsubscript{($0.67 \times$)} & 288k \textsubscript{($0.77 \times$)}  & 72.9 \textsubscript{(-0.4)}\\
\midrule
\midrule
\multirow{3}{*}{\bfseries MobileNetv3-Small} & \cmark & \xmark & \xmark & 14.5 \textsubscript{($1.00 \times$)} & 63k \textsubscript{($1.00 \times$)} & 188k \textsubscript{($1.00 \times$)} & 67.1 \textsubscript{(\textcolor{white}{+}0.0)} \\
 & \xmark & \cmark & \xmark & 24.7 \textsubscript{($1.70 \times$)}  & 71k \textsubscript{($1.13 \times$)} & 188k \textsubscript{($1.00 \times$)} & 66.8 \textsubscript{(-0.3)} \\
 & \xmark & \xmark & \cmark & 15.4 \textsubscript{($1.06 \times$)} & 57k \textsubscript{($0.90 \times$)}  & 144k \textsubscript{($0.77 \times$)} & 66.7 \textsubscript{(-0.4)}\\
\midrule
\multirow{3}{*}{\bfseries MobileNetv3-Large} & \cmark & \xmark & \xmark & 25.7         \textsubscript{($1.00 \times$)} & 71k \textsubscript{($1.00 \times$)} & 188k \textsubscript{($1.00 \times$)} & 75.3  \textsubscript{(\textcolor{white}{+}0.0)} \\
& \xmark & \cmark & \xmark & 47.3 \textsubscript{($1.84 \times$)} & 77k \textsubscript{($1.08 \times$)} & 188k \textsubscript{($1.00 \times$)} & 75.4 \textsubscript{(+0.1)} \\
& \xmark & \xmark & \cmark & 30.5 \textsubscript{($1.19 \times$)} & 60k \textsubscript{($0.85 \times$)} & 144k \textsubscript{($0.77 \times$)} & 75.1 \textsubscript{(-0.2)} \\
\bottomrule[1.5pt]
\end{tabular}
}
\caption{Effect of different samplers on the performance of different models.}
\label{tab:app_effect_samplers_diff_models}
\end{table*}

\end{document}